\documentclass[11pt]{article}
\usepackage{acl2016}
\usepackage{times}
\usepackage{url}
\usepackage{latexsym}

\usepackage{amsmath}
\usepackage{graphicx}

\aclfinalcopy 

\newcommand{\lden}{[\![}
\newcommand{\rden}{]\!]}
\newcommand{\denotes}[1]{\lden #1 \rden}
\DeclareMathOperator*{\argmax}{arg\,max}

\title{Resolving References to Objects in Photographs\\ using the
  Words-As-Classifiers Model}

\author{David Schlangen \;\; Sina Zarrie{\ss} \;\; Casey Kennington \\
  Dialogue Systems Group // CITEC // Faculty of Linguistics and Literary Studies\\
  Bielefeld University, Germany\\
  {\tt \emph{first.last}@uni-bielefeld.de}
}

\date{}

\begin{document}
\maketitle
\begin{abstract}
A common use of language is to refer to visually present objects. Modelling it in computers requires modelling the link between language and perception. 
The ``words as classifiers'' model of grounded semantics views words as classifiers of perceptual contexts, and composes the meaning of a phrase through composition of the denotations of its component words. 
It was recently shown to perform well in a game-playing scenario with a small number of object types. 
We apply it to two large sets of real-world photographs
that contain a much larger variety of object types and for which referring expressions are available. Using a pre-trained convolutional neural network to extract image region features, and augmenting these with positional information, we show that the model achieves performance competitive with the state of the art in a reference resolution task (\emph{given expression, find bounding box of its referent}), while, as we argue, being conceptually simpler and more flexible.
\end{abstract}

\section{Introduction}	

A common use of language is to refer to objects in the shared environment of speaker and addressee. Being able to simulate this is of particular importance for verbal human/robot interfaces  (\textsc{hri}), and the task has consequently received some attention in this field \cite{Matuszek2012,Tellex2011,Krishnamurthy2013}.

Here, we study a somewhat simpler precursor task, namely that of resolution of reference to objects in static images (photographs), but use a larger set of object types than is usually done in \textsc{hri} work ($>$ 300, see below). More formally, the task is to retrieve, given a referring expression $e$ and an image $I$, the region $bb^*$ of the image 
that is most likely to contain the referent of the expression.  As candidate regions, we use both manually annotated regions as well as automatically computed ones.



As our starting point, we use the ``words-as-classifiers'' model recently proposed by \newcite{kenschla:acl15}. 
It has before only been tested in a small domain and with specially designed features; here, we apply it to real-world photographs and use learned representations from a convolutional neural network \cite{googlenet}. We learn models for between 400 and 1,200 words, depending on the training data set.
As we show, the model performs competitive with the state of the art \cite{Huetal:saiaprref_Final,Maoetal:cocorefexp_Final} on the same data sets.

Our background interest in situated interaction makes it important for us that the approach we use is `dialogue ready'; and it is, in the sense that it supports incremental processing  (giving results while the incoming utterance is going on) and incremental learning (being able to improve performance from interactive feedback). However, in this paper we focus purely on `batch', non-interactive performance.\footnote{%
  The code for reproducing the results reported in this paper can be found at \url{https://github.com/dsg-bielefeld/image_wac}.
}

\section{Related Work}
\label{sec:relwo}

The idea of connecting words to what they denote in the real world via perceptual features goes back at least to \newcite{harnad:grounding}, who coined ``The Symbol Grounding Problem'': ``\emph{[H]ow can the semantic interpretation of a formal symbol system be made \emph{intrinsic} to the system, rather than just parasitic on the meanings in our heads?}'' The proposed solution was to 
link `categorial representations' with ``learned and innate feature detectors that pick out the invariant features of object and event categories from their sensory projections''. 

This suggestion has variously been taken up in computational work.
An early example is Deb Roy's work from the early 2000s \cite{royetal:visobsel,roy:visgrw,roy:trendsover}. In \cite{royetal:visobsel}, computer vision techniques are used to detect object boundaries in a video feed, and to compute colour features (mean colour pixel value), positional features, and features encoding the relative spatial configuration of objects. These features are then associated in a learning process with certain words, resulting in an association of colour features with colour words, spatial features with prepositions, etc., and based on this, these words can be interpreted with reference to the scene currently presented to the video feed. 

Of more recent work, that of \newcite{Matuszek2012} is closely related to the approach we take. The task in this work is to compute (sets of) referents, given a (depth) image of a scene containing simple geometric shapes and a natural language expression. In keeping with the formal semantics tradition, a layer of logical form representation is assumed; it is not constructed via syntactic parsing rules, however, but by a learned mapping (\emph{semantic parsing}). The non-logical constants of this representation then are interpreted by linking them to classifiers that work on perceptual features (representing shape and colour of objects). Interestingly, both mapping processes are trained jointly, and hence the links between classifiers and non-logical constants on the one hand, and non-logical constants and lexemes on the other are induced from data. In the work presented here, we take a simpler approach that forgoes the level of semantic representation and directly links lexemes and perceptions, but does not yet learn the composition. 

Most closely related on the formal side is recent work by \newcite{larsson:perc}, which offers a very direct implementation of the `words as classifiers' idea (couched in terms of type theory with records (\textsc{ttr}; \cite{coopginz:ttr-hbk}) and not model-theoretic semantics). In this approach, some lexical entries are enriched with classifiers that can judge, given a representation of an object, how applicable the term is to it. The paper also describes how these classifiers could be trained (or adapted) in interaction. The model is only specified theoretically, however, with hand-crafted classifiers for a small set of words, and not tested with real data. 

The second area to mention here is the recently very active one of image-to-text generation, which has been spurred on by the availability of large datasets and competitions structured around them.
The task here typically is to generate a description (a caption) for a given image. A frequently taken approach is to use a convolutional neural network (\textsc{cnn}) to map the image to a dense vector (which we do as well, as we will describe below), and then condition a neural language model (typically, an LSTM) on this to produce an output string \cite{vinyals:show,devlin:imcaqui}. 
\newcite{fangetal:2015} modify this approach somewhat, by using what they call ``word detectors'' first to specifically propose words for image regions, out of which the caption is then generated. This has some similarity to our word models as described below, but again is tailored more towards generation. 

\newcite{Socher2014} present a more compositional variant of this type of approach where sentence representations are composed along the dependency parse of the sentence. The representation of the root node is then mapped into a multimodal space in which distance between sentence and image representation can be used to guide image retrieval, which is the task in that paper. Our approach, in contrast, composes on the level of denotations and not that of representation.

Two very recent papers carry this type of approach over to the problem of resolving references to objects in images. Both \cite{Huetal:saiaprref} and \cite{Maoetal:cocorefexp} use \textsc{cnn}s to encode image information (and interestingly, both combine, in different ways, information from the candidate region with more global information about the image as a whole), on which they condition an \textsc{lstm} to get a prediction score for fit of candidate region and referring expression. As we will discuss below, our approach has some similarities, but can be seen as being more compositional, as the expression score is more clearly composed out of individual word scores (with rule-driven composition, however). We will directly compare our results to those reported in these papers, as we were able to use the same datasets.


\section{The ``Words-As-Classifiers'' Model}
\label{sec:wacform}

We now briefly review (and slightly reformulate) the model introduced by \newcite{kenschla:acl15}.
It has several components:

\paragraph{A Model of Word Meanings}

Let $w$ be a word whose meaning is to be modelled, and let $\mathbf{x}$ be a representation of an object in terms of its visual features. The core ingredient then is a classifier then takes this representation and returns a score $f_w(\mathbf{x})$, indicating the ``appropriateness'' of the word for denoting the object. 

Noting a (loose) correspondence to Montague's \shortcite{Tho:FP} intensional semantics, where the intension of a word is a function from possible worlds to extensions \cite{Gamut:2}, the \emph{intensional} meaning of $w$ is then defined as the classifier itself, a function from a representation of an object to an ``appropriateness score'':\footnote{%
  \cite{larsson:perc} develops this intension/extension distinction in more detail for his formalisation.
}

\begin{center}
{\footnotesize
\vspace*{-0.6cm}
\begin{equation}
  \denotes{w}_{obj} = \lambda \mathbf{x}. f_w(\mathbf{x})
\label{eq:intensobj}
\end{equation}
}
\end{center}
\vspace*{-0.2cm}

(Where $\denotes{.}$ is a function returning the meaning of its argument, and $\mathbf{x}$ is a feature vecture as given by  $f_{obj}$, the function that computes the representation for a given object.)

The \emph{extension} of a word in a given (here, visual) discourse universe $W$ can then be modelled as a probability distribution ranging over all candidate objects in the given domain, resulting from the application of the word intension to each object ($\mathbf{x}_i$ is the feature vector for object $i$, $normalize()$ vectorized normalisation, and $I$ a random variable ranging over the $k$ candidates):

\vspace{-0.7cm}
\begin{center}
{\footnotesize
\begin{multline}
   \denotes{w}^W_{obj} = \\
   normalize((\denotes{w}_{obj}(\mathbf{x}_1), \dots, \denotes{w}_{obj}(\mathbf{x}_k))) = \\
   normalize((f_w(\mathbf{x}_1), \dots, f_w(\mathbf{x}_k))) = P(I|w)
\label{eq:word}
\end{multline}
}
\end{center}

\paragraph{Composition}

Composition of word meanings into phrase meanings in this approach is governed by rules that are tied to syntactic constructions. In the following, we only use simple multiplicative composition for nominal constructions: 

\vspace{-0.8cm}
\begin{center}
{\footnotesize
\begin{multline}
\denotes{[_{nom}w_1,\dots,w_k]}^W = \denotes{\textsc{nom}}^W\denotes{w_1,\dots,w_k}^W =\\ \circ_{/N}(\denotes{w_1}^W,\dots,\denotes{w_k}^W)
\label{eq:re}
\end{multline}
}
\end{center}
\vspace{-0.5cm}
where $\circ_{/N}$ is defined as

\vspace{-0.6cm}
{\footnotesize
\begin{multline}
   \circ_{/N}(\denotes{w_1}^W, \dots, \denotes{w_k}^W) = P_{\circ}(I|w_1,\dots,w_k) \\
   \mbox{\ with\ } 
    P_{\circ}(I=i|w_1,\dots, w_k) = \\
    \frac{1}{Z} (P(I=i|w_1) * \dots * P(I=i|w_k))  \mbox{ for } i\in I
\label{eq:avg}
\end{multline}
}
\vspace{-0.55cm}

\noindent
($Z$ takes care that the result is normalized over all candidate objects.)

\paragraph{Selection}

To arrive at the desired extension of a full referring expression---an individual object, in our case---, one additional element is needed, and this is contributed by the determiner. For uniquely referring expressions (``the red cross''), what is required is to pick the most likely candidate from the distribution:

\vspace{-0.6cm}
{\footnotesize
\begin{equation}
\denotes{the} = \lambda x. \argmax_{Dom(x)} x
\label{eq:the}
\end{equation}
}

\begin{center}
{\footnotesize
\vspace{-1.2cm}
\begin{multline}
\denotes{[the]\ [_{nom}w_1,\dots,w_k]}^W = \\
\argmax_{i\in W} [\; \denotes{[_{nom}w_1,\dots,w_k]}^W \;]
\label{eq:thephr}
\end{multline}
}
\end{center}

\begin{figure}[ht]
  \hspace*{-.3cm}
  \includegraphics[width=1.15\columnwidth]{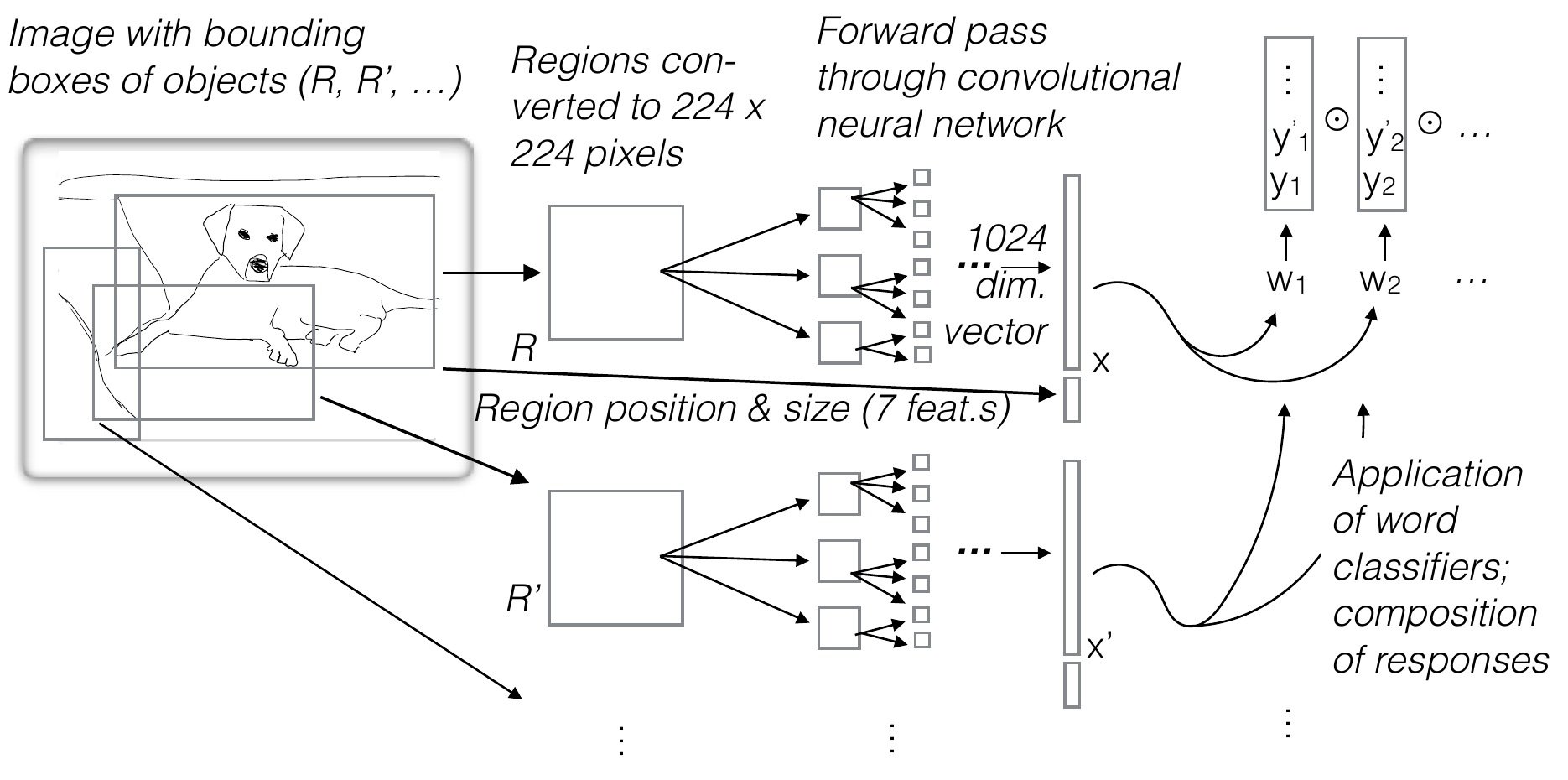}
  \vspace*{-.8cm}
  \caption{Overview of the model}
  \label{fig:model}
\end{figure}

\noindent 
In other words, the prediction of an expression such as ``the brown shirt guy on right'' is computed by first getting the responses of the classifiers corresponding to the words, individually for each object. I.e., the classifier for ``brown'' is applied to objects $o_1$, \dots, $o_n$. This yields a vector of responses (of dimensionality $n$, the number of candidate objects); similarly for all other words. 
These vectors are then multiplied, and the predicted object is the maximal component of the resulting vector. Figure~\ref{fig:model} gives a schematic overview of the model as implemented here, including the feature extraction process.

\section{Data: Images \& Referring Expressions}
\label{sec:data}


\begin{figure*}[ht]
  \centering
      \includegraphics[width=0.25\textwidth]{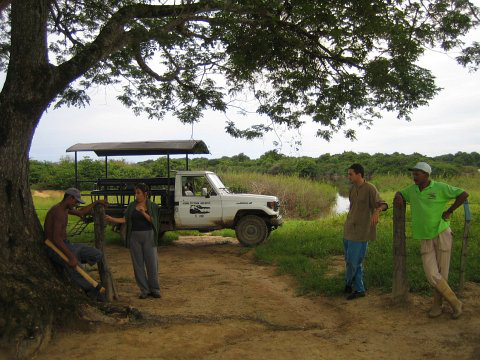}	
      \;
      \includegraphics[width=0.25\textwidth]{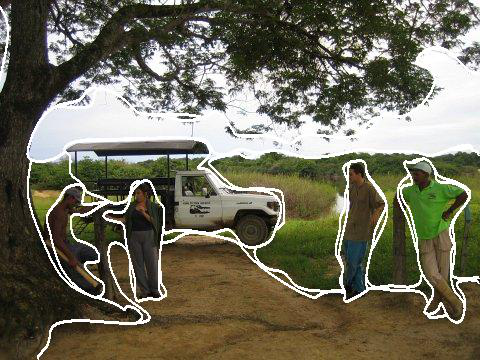}	
      \;
      \includegraphics[width=0.25\textwidth]{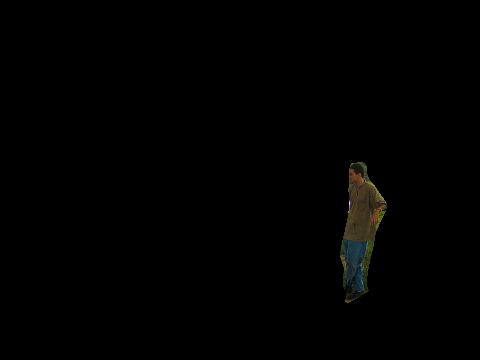}	
      \caption{Image 27437 from \textsc{iapr tc-12} (left), with region masks from \textsc{saiapr tc-12} (middle); ``brown shirt guy on right'' is a referring expression in \textsc{ReferItGame} for the region singled out on the right }
      \label{fig:exim}
\end{figure*}

\paragraph{SAIAPR TC-12 / ReferItGame} The basis of this data set is the \textsc{iapr tc-12} image retrieval benchmark collection of ``20,000 still natural images taken from locations around the world and comprising an assorted cross-section of still natural images'' \cite{Grubinger2006}. A typical example of an image from the collection is shown in Figure~\ref{fig:exim} on the left. 

This dataset was later  augmented by \newcite{Escalante2010} with segmentation masks identifying objects in the images (an average of 5 objects per image). Figure~\ref{fig:exim} (middle) gives an example of such a segmentation. These segmentations were done manually and provide close maskings of the objects. 
This extended dataset is also known as ``\textsc{saiapr tc-12}'' (for ``segmented and annotated \textsc{iapr tc-12}'').

The third component is provided by \newcite{Kazemzadeh2014}, who collected a large number of expressions referring to (pre-segmented) objects from these images, using a crowd-sourcing approach where two players were paired and a director needed to refer to a predetermined object to a matcher, who then selected it.
(An example is given in Figure~\ref{fig:exim} (right).)
This corpus contains 120k referring expressions, covering nearly all of the 99.5k regions from \textsc{saiapr tc-12}.\footnote{The \textsc{iapr tc-12} and \textsc{saiapr tc-12} data is available from \url{http://imageclef.org}; \textsc{ReferItGame} from \url{http://tamaraberg.com/referitgame}.} The average length of a referring expression from this corpus is 3.4 tokens. The 500k token realise 10,340 types, with 5785 hapax legomena. The most frequent tokens (other than articles and prepositions) are ``left'' and ``right'', with 22k occurrences.
(In the following, we will refer to this corpus as \textsc{referit}.)

This combination of segmented images and referring expressions has recently been used by \newcite{Huetal:saiaprref} for learning to resolve references, as we do here. The authors also tested their method on region proposals computed using the EdgeBox algorithm \cite{zitnick2014edge}. They kindly provided us with this region proposal data (100 best proposals per image), and we compare our results on these region proposals with theirs below. The authors split the dataset evenly into 10k images (and their corresponding referring expressions) for training and 10k for testing. As we needed more training data, we made a 90/10 split, ensuring that all our test images are from their test split.

\paragraph{MSCOCO / GoogleRefExp / ReferItGame}

\begin{figure}[ht]
  \begin{minipage}[c]{.7\columnwidth}
    \includegraphics[width=\columnwidth]{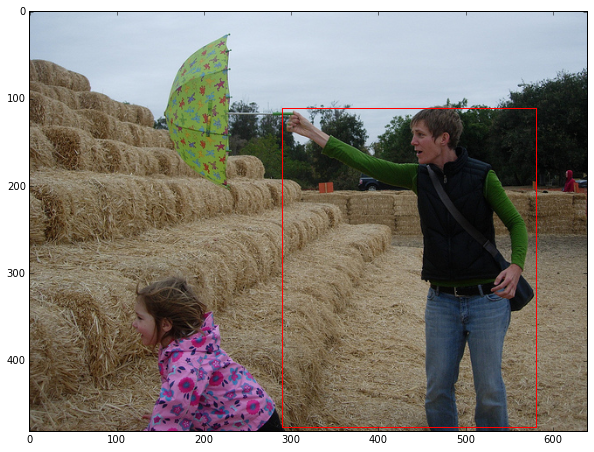}
  \end{minipage}
  \begin{minipage}[t]{.29\columnwidth}
  \vspace*{-1cm}
    {\small
      \textsc{refcoco:}\\ green woman\\
      \textsc{grexp:}\\ a woman wearing blue jeans
    }
  \end{minipage}
  \begin{minipage}[c]{.7\columnwidth}
    \includegraphics[width=\columnwidth]{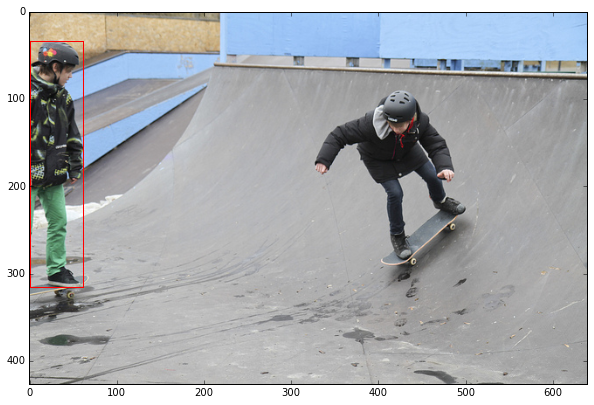}
  \end{minipage}
  \begin{minipage}[t]{.29\columnwidth}
  \vspace*{-1.5cm}
    {\small
      \textsc{refcoco:}\\ person left\\
      \textsc{grexp:}\\ a boy is ready to play who is wearing green color pant
    }
  \end{minipage}
  \caption{Examples from \textsc{mscoco}}
\vspace*{-.6cm}
  \label{fig:coco}
\end{figure}

The second dataset is based on the ``Microsoft Common Objects in Context'' collection \cite{mscoco}, which contains over 300k images with object segmentations (of objects from 80 pre-specified categories), object labels, and image captions. Figure~\ref{fig:coco} shows some examples of images containing objects of type ``person''.

This dataset was augmented by \newcite{Maoetal:cocorefexp} with what they call `unambiguous object descriptions', using a subset of 27k images that contained between 2 and 4 instances of the same object type within the same image. The authors collected and validated 100k descriptions in a crowd-sourced approach as well, but unlike in the ReferItGame setup, describers and validators were not connected live in a game setting.\footnote{%
  The data is available from \url{https://github.com/mjhucla/Google_Refexp_toolbox}.
}
The average length of the descriptions is 8.3 token. The 790k token in the corpus realise 14k types, with 6950 hapax legomena. The most frequent tokens other than articles and prepositions are ``man'' (15k occurrences) and ``white'' (12k). (In the following, we will refer to this corpus as \textsc{grexp}.)

The authors also computed automatic region proposals for these images, using the multibox method of \newcite{Erhan2014} and classifying those using a model trained on \textsc{mscoco} categories, retaining on average only 8 per image. These region proposals are on average of a much higher quality than those we have available for the other dataset.

As mentioned in \cite{Maoetal:cocorefexp}, Tamara Berg and colleagues have at the same time used their ReferItGame paradigm to collect referring expressions for \textsc{mscoco} images as well. Upon request, Berg and colleagues also kindly provided us with this data---140k referring expressions, for 20k images, average length 3.5 token, 500k token altogether, 10.3k types, 5785 hapax legomena; most frequent also ``left'' (33k occurrences) and ``right'' (32k). (In the following, we will call this corpus \textsc{refcoco}.) In our experiments, we use the training/validation/test splits on the images suggested by Berg et al., as the splits provided by \newcite{Maoetal:cocorefexp} are on the level of objects and have some overlap in images. 

It is interesting to note the differences in the expressions from \textsc{refcoco} and \textsc{grexp}, the latter on average being almost 5 token longer. Figure~\ref{fig:coco} gives representative examples. We can speculate that the different task descriptions (``refer to this object'' vs.\ ``produce an unambiguous description'') and the different settings (live to a partner vs.\ offline, only validated later) may have caused this. As we will see below, the \textsc{grexp} descriptions did indeed cause more problems to our approach, which is meant for reference in interaction.

\section{Training the Word/Object Classifiers}
\label{sec:training}

The basis of the approach we use are the classifiers that link words and images. These need to be trained from data; more specifically, from pairings of image regions and referring expressions, as provided by the corpora described in the previous section.

\paragraph{Representing Image Regions} The first step is to represent the information from the image regions. We use a deep convolutional neural network, ``GoogLeNet'' \cite{googlenet}, that was trained on data from the Large Scale Visual Recognition Challenge 2014 (ILSVRC2014) from the ImageNet corpus \cite{imagenet_cvpr09} to extract features.\footnote{%
  \url{http://www.image-net.org/challenges/LSVRC/2014/}.\\
  We use the sklearn-theano (\url{http://sklearn-theano.github.io/feature_extraction/index.html#feature-extraction}) port of the Caffe replication and re-training  (\url{https://github.com/BVLC/caffe/tree/master/models/bvlc_googlenet}) of this network structure.}
It was optimised to recognise categories from that challenge, which are different from those occurring in either \textsc{saiapr} or \textsc{coco}, but in any case we only use the final fully-connected layer before the classification layer, to give us a 1024 dimensional representation of the region. We augment this with 7 features that encode information about the region relative to the image: the (relative) coordinates of two corners, its (relative) area, distance to the center, and orientation of the image. The full representation hence is a vector of 1031 features. (See also Figure~\ref{fig:model} above.)

\paragraph{Selecting candidate words}

How do we select the words for which we train perceptual classifiers?
There is a technical consideration to be made here and a semantic one. The technical consideration is that we need sufficient training data for the classifiers, and so can only practically train classifiers for words that occur often enough in the training corpus. We set a threshold here of a minimum of 40 occurences in the training corpus, determined empirically on the validation set to provide a good tradeoff between vocabulary coverage and number of training instances.


The semantic consideration is that intuitively, the approach does not seem appropriate for all types of words; where it might make sense for attributes and category names to be modelled as image classifiers, it does less so for prepositions and other function words. Nevertheless, for now, we make the assumption 
that all words in a referring expression contribute information to the \emph{visual} identification of its referent. We discuss the consequences of this decision below.

This assumption is violated in a different way in phrases that refer via a landmark, such as in ``the thing next to the woman with the blue shirt''. 
Here we cannot assume for example that the referent region provides a good instance of ``blue'' (since it is not the target object in the region that is described as blue), and so we exclude such phrases from the training set (by looking for a small set of expressions such as ``left of'', ``behind'', etc.; see appendix for a full list). This reduces the training portions of \textsc{referit}, \textsc{refcoco} and \textsc{grexp} to 86\%, 95\%, and 82\% of their original size, respectively (counting referring expressions, not tokens).

Now that we have decided on the set of words for which to train classifiers, how do we assemble the training data? 

\paragraph{Positive Instances}

Getting positive instances from the corpus is straightforward: We pair each word in a referring expression with the representation of the region it refers to. That is, if the word ``left'' occurs 20,000 times in expressions in the training corpus, we have 20,000 positive instances for training its classifier.

\paragraph{Negative Instances}

Acquiring negative instances is less straightforward. The corpus does not record inappropriate uses of a word, or `negative referring expressions' (as in ``this is not a red chair''). To create negative instances, we make a second assumption which again is not generally correct, namely that when a word was \emph{never} in the corpus used to refer to an object, this object can serve as a negative example for that word/object classifier. In the experiments reported below, we randomly selected 5 image regions from the training corpus whose referring expressions (if there were any) did not contain the word in question.\footnote{%
  This approach is inspired by the negative sampling technique of \newcite{Mikolov2013:embeddings} for training textual word embeddings.
}

\paragraph{The classifiers}

Following this regime, we train binary logistic regression classifiers (with $\ell 1$ regularisation) on the visual object features representations, for all words that occurred at least 40 times in the respective training corpus.\footnote{%
  We used the implementation in the \texttt{scikit learn} package \cite{scikit-learn}.
}

\vspace*{.2cm} \noindent
To summarise, we train separate binary classifiers for each word (not making any a-priori distinction between function words and others, or attribute labels and category labels), giving them the task to predict how likely it would be that the word they represent would be used to refer to the image region they are given. 
All classifiers are presented during training with data sets with the same balance of positive and negative examples (here, a fixed ratio of 1 positive to 5 negative). Hence, the classifiers themselves do not reflect any word frequency effects; our claim (to be validated in future work) is that any potential effects of this type are better modelled separately.

\section{Experiments}
\label{sec:exp}

The task in our experiments is the following: Given an image $I$ together with bounding boxes of regions ($bb_1, \dots, bb_n$) within it, and a referring expression $e$, predict which of these regions contains the referent of the expression. 


\paragraph{By Corpus}
\label{sec:bycorp}


We start with training and testing models for all three corpora (\textsc{referit}, \textsc{refcoco}, \textsc{grexp}) separately. But first, we establish some baselines. 
The first is just randomly picking one of the candidate regions. The second is a 1-rule classifier that picks the largest region. The respective accuracies on the corpora are as follows: \textsc{referit} 0.20/0.19; \textsc{refcoco} 0.16/0.23; \textsc{grexp} 0.19/0.20.

\begin{table}
\centering
{\small
\hspace*{-.5cm}
\begin{tabular}{l||rrrr|rrrr}
{} &  \%tst &  acc &  mrr &  arc &  $>$0 &  acc \\
\hline
\hline
\textsc{referit}     &  1.00 &      0.65 &      0.79 &      0.89 &      0.97 &      0.67 \\
\textsc{referit}; NR &  0.86 &      0.68 &      0.82 &      0.91 &      0.97 &      \textbf{0.71} \\
\cite{Huetal:saiaprref} & -- & 0.73 & -- & -- & -- & -- \\
\hline
\textsc{refcoco}   &  1.00 &      0.61 &      0.77 &      0.91 &      0.98 &      0.62 \\
\textsc{refcoco}; NR &  0.94 &      0.63 &      0.78 &      0.92 &      0.98 &      \textbf{0.64} \\
\cite{Maoetal:cocorefexp} & -- & 0.70 & -- & -- & -- & -- \\
\hline
\textsc{grexp}     &  1.00 &      0.43 &      0.65 &      0.86 &      1.00 &      0.43 \\
\textsc{grexp}; NR &  0.82 &      0.45 &      0.67 &      0.88 &      1.00 &      \textbf{0.45} \\
\cite{Maoetal:cocorefexp} & -- & 0.61 & -- & -- & -- & -- \\
\end{tabular}
}
\vspace*{-.5cm}
\caption{Results; separately by corpus. See text for description of columns and rows.}
\vspace*{-.5cm}
\label{tab:res1}
\end{table}

Training on the training sets of \textsc{referit}, \textsc{refcoco} and \textsc{grex} with the regime described above (min.\ 40 occurrences) gives us classifiers for 429, 503, and 682 words, respectively. Table~\ref{tab:res1} shows the evaluation on the respective test parts: accuracy (\emph{acc}), mean reciprocal rank (\emph{mrr}) and for how much of the expression, on average, a word classifier is present (\emph{arc}). `$>$0' shows how much of the testcorpus is left if expressions are filtered out for which not even a single word is the model (which we evaluate by default as false), and accuracy for that reduced set. The `NR' rows give the same numbers for reduced test sets in which all relational expressions have been removed; `\%tst' shows how much of a reduction that is relative to the full testset. 
The rows with the citations give the best reported results from the literature.\footnote{%
  Using a different split than \cite{Maoetal:cocorefexp}, as their train/test set overlaps on the level of images.
}

 As this shows, in most cases we come close, but do not quite reach these results. The distance is the biggest for \textsc{grexp} with its much longer expressions. As discussed above, not only are the descriptions longer on average in this corpus, the vocabulary size is also much higher. Many of the descriptions contain action descriptions (``the man smiling at the woman''), which do not seem to be as helpful to our model. Overall, the expressions in this corpus do appear to be more like `mini-captions' describing the region rather than referring expressions that efficiently single it out among the set of distractors; our model tries to capture the latter.


\paragraph{Combining Corpora}
\label{sec:comb}

A nice effect of our setup is that we can freely mix the corpora for training, as image regions are represented in the same way regardless of source corpus, and we can combine occurrences of a word across corpora. We tested combining the testsets of \textsc{referit} and \textsc{refcoco} (\textsc{ri+rc} in the Table below), \textsc{refcoco} and \textsc{grexp} (\textsc{rc+gr}), and all three (\textsc{referit}, \textsc{refcoco}, and \textsc{grexp}; \textsc{ri+rc+gr}), yielding models for 793, 933, 1215 words, respectively (with the same ``min.\ 40 occurrences'' criterion). For all testsets, the results were at least stable compared to Table~\ref{tab:res1}, for some they improved. For reasons of space, we only show the improvements here.

\begin{table}[h]
  \centering
{\small
\hspace*{-.8cm}
\begin{tabular}{lrrrr|rr}
{} &  \%tst &  acc &  mrr &  arc &  $>$0 &  acc \\
\hline
\hline
\textsc{ri+rc}/\textsc{rc}      &  1.00 &      \textbf{0.63} &      0.78 &      0.92 &      0.98 &      \textbf{0.64}\\
\textsc{ri+rc}/\textsc{rc}; NR  &  0.94 &      \textbf{0.65} &      0.79 &      0.93 &      0.98 &      \textbf{0.66}\\
\hline
\textsc{ri+rc+gr}/\textsc{rc}     &  1.00 &      \textbf{0.63} &      0.78 &      0.94 &      0.99 &      \textbf{0.64}  \\
\textsc{ri+rc+gr}/\textsc{rc}; NR &  0.94 &      \textbf{0.65} &      0.79 &      0.95 &      0.99 &      \textbf{0.66} \\
\hline
\textsc{ri+rc+gr}/\textsc{gr}     &  1.00 &      \textbf{0.47} &      0.68 &      0.90 &      1.00 &      \textbf{0.47}  \\
\textsc{ri+rc+gr}/\textsc{gr}; NR &  0.82 &      \textbf{0.49} &      0.70 &      0.91 &      1.00 &      \textbf{0.49}\\
\end{tabular}
}
\vspace*{-.5cm}
  \caption{Results, combined corpora}
  \label{tab:rescomb}
\end{table}




\paragraph{Computed Region Proposals}
\label{sec:rprops}
Here, we cannot expect the system to retrieve exactly the ground truth bounding box, since we cannot expect the set of automatically computed regions to contain it. We follow \newcite{Maoetal:cocorefexp} in using \emph{intersection over union} (IoU) as metric (the size of the intersective area between candidate and ground truth bounding box normalised by the size of the union) and taking an IoU $\ge$ 0.5 of the top candidate as a threshold for success (P@1). As a more relaxed metric, we also count for the \textsc{saiapr} proposals (of which there are 100 per image) as success when at least one among the top 10 candidates exceeds this IoU threshold (R@10). (For \textsc{mscoco}, there are only slightly above 5 proposals per image on average, so computing this more relaxed measure does not make sense.) The random baseline (\textsc{rnd}) is computed by applying the P@1 criterion to a randomly picked region proposal. (That it is higher than 1/\#regions for \textsc{saiapr} shows that the regions cluster around objects.)

\begin{table}[h]
  \centering
{\small
\begin{tabular}{lrrr}
{} &  RP@1 &  RP@10 &  rnd \\
\hline
\hline
\textsc{referit}         &       0.09 &        0.24 &      0.03 \\
\textsc{referit}; NR &       \textbf{0.10} &        0.26 &      0.03 \\
\cite{Huetal:saiaprref} & 0.18 & 0.45 \\
\hline
\textsc{refcoco}        &       0.52 &       -- &      0.17 \\
\textsc{refcoco}; NR &       \textbf{0.54} &        -- &      0.17 \\
\cite{Maoetal:cocorefexp} & 0.52\\
\hline
\textsc{grexp}         &       0.36 &        -- &      0.16 \\
\textsc{grexp}; NR &       \textbf{0.37} &         -- &      0.17 \\
\cite{Maoetal:cocorefexp} & 0.45\\
\end{tabular}
}
  \caption{Results on region proposals}
  \label{tab:rprops}
\end{table}

With the higher quality proposals provided for the \textsc{mscoco} data, and the shorter, more prototypical referring expressions from \textsc{refcoco}, we narrowly beat the reported results. (Again, note that we use a different split that ensures separation on the level of images between training and test.)
\cite{Huetal:saiaprref} performs relatively better on the region proposals (the gap is wider), on \textsc{grexp}, we come relatively closer using these proposals.
We can speculate that using automatically computed boxes of a lower selectivity (\textsc{referit}) shifts the balance between needing to actually recognise the image and getting information from the shape and position of the box (our \emph{positional} features; see Section~\ref{sec:training}).


\paragraph{Ablation Experiments}
\label{sec:abl}

To get an idea about what the classifiers actually pick up on, we trained variants given only the positional features (\textsc{pos} columns below in Table~\ref{tab:abla}) and only object features (\textsc{nopos} columns). We also applied a variant of the model with only the top 20 classifiers (in terms of number of positive training examples; \textsc{top20}). We only show accuracy here, and repeat the relevant numbers from Table~\ref{tab:res1} for comparison (\textsc{full}).

\begin{table}[h]
  \centering
{\small
\begin{tabular}{l|rr|r|r}
{} & nopos &  pos &  full & top20\\
\hline
\hline
\textsc{ri}     & 0.53 & 0.60 & 0.65 & 0.46\\
\textsc{ri}; NR & 0.56 & 0.62 & 0.68 & 0.48\\
\hline
\textsc{rc}     & 0.44 & 0.55 & 0.61 & 0.52\\
\textsc{rc}; NR & 0.45 & 0.57 & 0.63 & 0.53
\end{tabular}
}
  \caption{Results with reduced models}
  \label{tab:abla}
\end{table}


This table shows an interesting pattern. To a large extent, the object image features and the positional features seem to carry redundant information, with the latter on their own performing better than the former on their own. The full model, however, still gains something from the combination of the feature types. The top-20 classifiers (and consequently, top 20 most frequent words) alone reach decent performance (the numbers are shown for the full test set here; if reduced to only utterances where at least one word is known, the numbers rise, but the reduction of the testset is much more severe than for the full models with much larger vocabulary).

\paragraph{Error Analysis}
\label{sec:errors}


\begin{figure}[ht]
  \centering
  \includegraphics[width=\columnwidth]{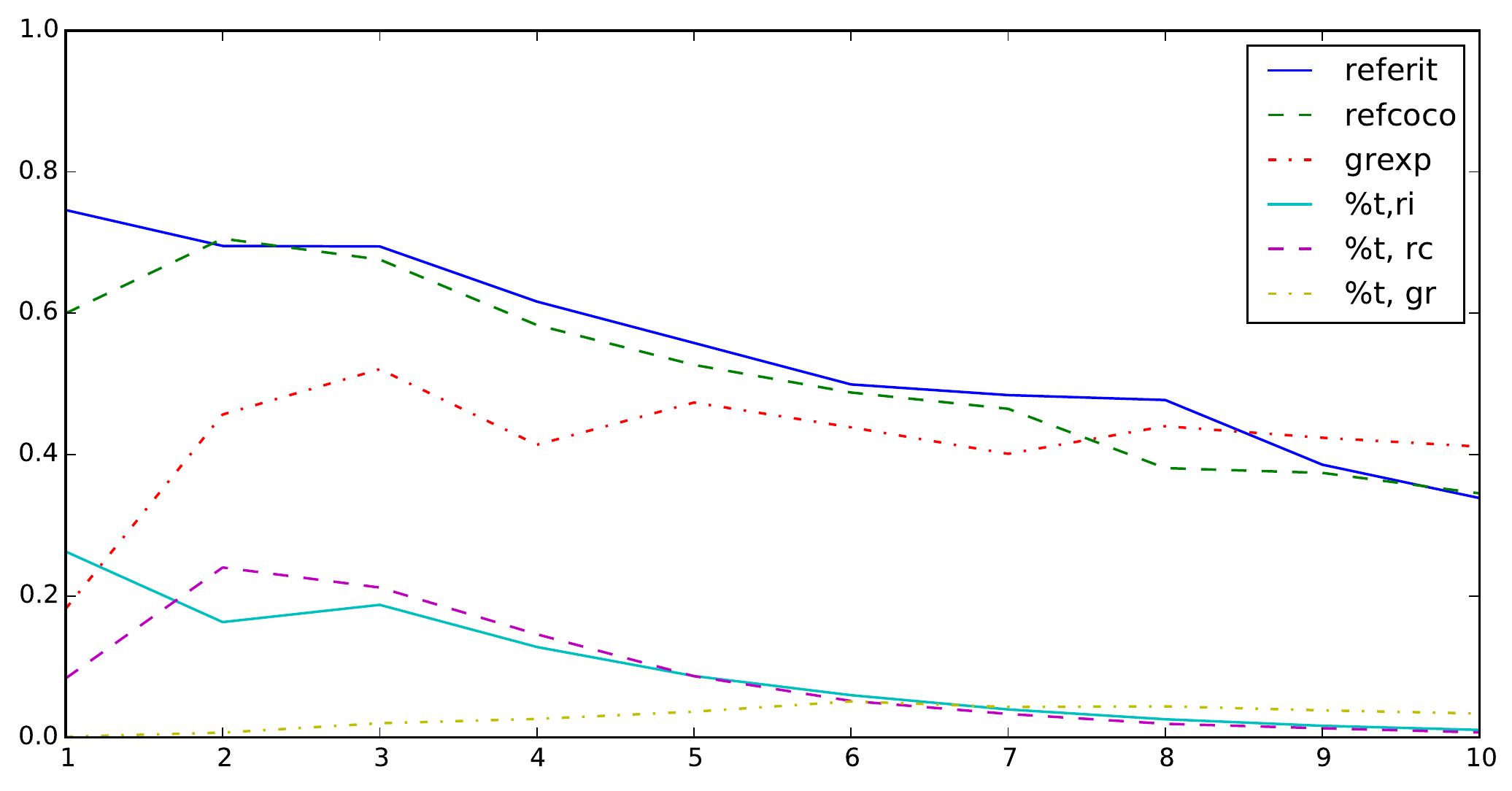}
  \vspace*{-.7cm}
  \caption{Accuracy by expression length (top 3 lines); percentage of expressions with this length (lower 3 lines).}
  \label{fig:bylength}
\end{figure}

Figure~\ref{fig:bylength} shows the accuracy of the model split by length of the referring expression (top lines; lower lines show the proportion of expression of this length in the whole corpus). The pattern is similar for all corpora (but less pronounced for \textsc{grexp}): shorter utterances fare better. 


Manual inspection of the errors made by the system further corroborates the suspicion that composition as done here neglects too much of the internal structure of the expression. An example from \textsc{referit} where we get a wrong prediction is ``second person from left''. The model clearly does not have a notion of counting, and here it wrongly selects the leftmost person. In a similar vein, we gave results above for a testset where spatial relations where removed, but other forms of relation (e.g., ``child sitting on womans lap'') that weren't modelled still remain in the corpus.  

\begin{figure}[ht]
  \centering
  \includegraphics[width=.8\columnwidth]{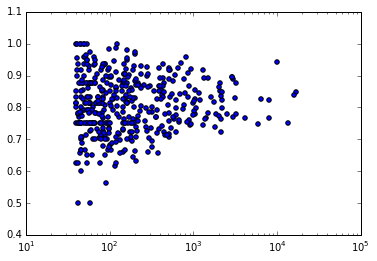}
  \vspace*{-.3cm}
  \caption{Average accuracy vs.\ \# train instanc.}
  \label{fig:tva}
\end{figure}

We see as an advantage of the model that we can inspect words individually. Given the performance of short utterances, we can conclude that the word/object classifiers themselves perform reasonably well. This seems to be somewhat independent of the number of training examples they received. Figure~\ref{fig:tva} shows, for \textsc{referit}, \# training instances (x-axis) vs.\ average accuracy on the validation set, for the whole vocabulary. As this shows, the classifiers tend to get better with more training instances, but there are good ones even with very little training material.

Mean average precision (i.e., area under the precision / recall curve) over all classifiers (exemplarily computed for the \textsc{ri+rc} set, 793 words) is $0.73$ (std $0.15$). Interestingly, the 155 classifiers in the top range (average precision over $0.85$) are almost all for concrete nouns; the 128 worst performing ones (below $0.60$) are mostly other parts of speech. (See appendix.) This is, to a degree, as expected: our assumption behind training classifiers for \emph{all} ocurring words and not pre-filtering based on their part-of-speech or prior hypotheses about visual relevance was that words that can occur in all kinds of visual contexts will lead to classifiers whose contributions cancel out across all candidate objects in a scene.

However, the mean average precision of the classifiers for colour words is also relatively low at $0.6$ (std $0.08$), for positional words (``left'', ``right'', ``center'', etc.) it is $0.54$ (std $0.1$). This might suggest that the features we take from the CNN  might indeed be more appropriate for tasks close to what they were originally trained on, namely category and not attribute prediction. We will explore this in future work.




\section{Conclusions}
\label{sec:conc}

We have shown that the ``words-as-classifiers'' model scales up to a larger set of object types with a much larger variety in appearance (\textsc{saiapr} and \textsc{mscoco}); to a larger vocabulary and much less restricted expressions (\textsc{referit}, \textsc{refcoco}, \textsc{grexp}); and to use of automatically learned feature types (from a \textsc{cnn}). It achieves results that are comparable to those of more complex models. 

We see as advantage that the model we use is ``transparent'' and modular. Its basis, the word/object classifiers, ties in more directly with more standard approaches to semantic analysis and composition. Here, we have disregarded much of the internal structure of the expressions. But there is a clear path for bringing it back in, by defining other composition types for other construction types and different word models for other word types. \newcite{kenschla:acl15} do this for spatial relations in their simpler domain; for our domain, new and more richly annotated data such as \textsc{visual}genome looks promising for learning a wide variety of relations.\footnote{%
\url{http://visualgenome.org/}} 
The use of denotations / extensions might make possible transfer of methods from extensional semantics, e.g.\ for the addition of operators such as negation or generalised quantifiers. 
The design of the model, as mentioned in the introduction, makes it amenable for use in interactive systems that learn; we are currently exploring this avenue. Lastly, the word/object classifiers also show promise in the reverse task, generation of referring expressions \cite{zarrieschlang:easy-pre}.

All this is future work. In its current state---besides, we believe, strongly motivating this future work---, we hope that the model can also serve as a strong baseline to other future approaches to reference resolution, as it is conceptually simple and easy to implement. 









\section*{Acknowledgments}

We thank \newcite{Huetal:saiaprref_Final}, \newcite{Maoetal:cocorefexp_Final} and Tamara Berg for giving us access to their data. Thanks are also due to the anonymous reviewers for their very insightful comments. We acknowledge support by the Cluster of Excellence ``Cognitive Interaction Technology'' (CITEC; EXC 277) at Bielefeld University, which is funded by the German Research Foundation (DFG), and by the DUEL project, also funded by DFG (grant SCHL 845/5-1).




\bibliography{refs_acl16}
\bibliographystyle{acl2016}

\appendix

\section{Supplemental Material}
\label{sec:supplemental}

\paragraph{Filtering relational expressions}

As described above, we filter out all referring expressions during training that contain either of the following tokens:

{\small
\begin{verbatim}
RELWORDS = ['below',
            'above',
            'between',
            'not',
            'behind',
            'under',
            'underneath',
            'front of',
            'right of',
            'left of',
            'ontop of',
            'next to',
            'middle of']
\end{verbatim}
}

\paragraph{Average Precision}

See Section~\ref{sec:errors}. 

\noindent
Classifiers with average precision over $0.85$:
{\small
\begin{verbatim}
['giraffe', 'coffee', 'court', 'riding', 'penguin', 'balloon', 'ball',
'mug', 'turtle', 'tennis', 'beer', 'seal', 'cow', 'bird', 'horse',
'drink', 'koala', 'sheep', 'ceiling', 'parrot', 'bike', 'cactus',
'sun', 'smoke', 'llama', 'fruit', 'ruins', 'waterfall', 'nightstand',
'books', 'night', 'coke', 'skirt', 'leaf', 'wheel', 'label', 'pot',
'animals', 'cup', 'tablecloth', 'pillar', 'flag', 'field', 'monkey',
'bowl', 'curtain', 'plate', 'van', 'surfboard', 'bottle', 'fish',
'umbrella', 'bus', 'shirtless', 'train', 'bed', 'painting', 'lamp',
'metal', 'paper', 'sky', 'luggage', 'player', 'face', 'going', 'desk',
'ship', 'raft', 'lying', 'vehicle', 'trunk', 'couch', 'palm', 'dress',
'doors', 'fountain', 'column', 'cars', 'flowers', 'tire', 'plane',
'against', 'bunch', 'car', 'shelf', 'bunk', 'boat', 'dog', 'vase',
'animal', 'pack', 'anyone', 'clock', 'glass', 'tile', 'window',
'chair', 'phone', 'across', 'cake', 'branches', 'bicycle', 'snow',
'windows', 'book', 'curtains', 'bear', 'guitar', 'dish', 'both',
'tower', 'truck', 'bridge', 'creepy', 'cloud', 'suit', 'stool', 'tv',
'flower', 'seat', 'buildings', 'shoes', 'bread', 'hut', 'donkey',
'had', 'were', 'fire', 'food', 'turned', 'mountains', 'city', 'range',
'inside', 'carpet', 'beach', 'walls', 'ice', 'crowd', 'mirror',
'brush', 'road', 'anything', 'blanket', 'clouds', 'island',
'building', 'door', '4th', 'stripes', 'bottles', 'cross', 'gold',
'smiling', 'pillow']
\end{verbatim}
}

\noindent
Classifiers with average precision below $0.6$:
{\small
\begin{verbatim}
['shadow', "woman's", 'was', 'bright', 'lol', 'blue', 'her', 'yes',
'blk', 'this', 'from', 'almost', 'colored', 'looking', 'lighter',
'far', 'foreground', 'yellow', 'looks', 'very', 'second', 'its',
'dat', 'stack', 'dudes', 'men', 'him', 'arm', 'smaller', 'half',
'piece', 'out', 'item', 'line', 'stuff', 'he', 'spot', 'green',
'head', 'see', 'be', 'black', 'think', 'leg', 'way', 'women',
'furthest', 'rt', 'most', 'big', 'grey', 'only', 'like', 'corner',
'picture', 'shoulder', 'no', 'spiders', 'n', 'has', 'his', 'we',
'bit', 'spider', 'guys', '2', 'portion', 'are', 'section', 'us',
'towards', 'sorry', 'where', 'small', 'gray', 'image', 'but',
'something', 'center', 'i', 'closest', 'first', 'middle', 'those',
'edge', 'there', 'or', 'white', '-', 'little', 'them', 'barely',
'brown', 'all', 'mid', 'is', 'thing', 'dark', 'by', 'back', 'with',
'other', 'near', 'two', 'screen', 'so', 'front', 'you', 'photo', 'up',
'one', 'it', 'space', 'okay', 'side', 'click', 'part', 'pic', 'at',
'that', 'area', 'directly', 'in', 'on', 'and', 'to', 'just', 'of']
\end{verbatim}
}

\paragraph{Code}

The code required for reproducing the results reported here can be found at \url{https://github.com/dsg-bielefeld/image_wac}.






\end{document}